\newcommand{\CLASSINPUTtoptextmargin}{0.75in}
\newcommand{\CLASSINPUTbottomtextmargin}{0.75in}
\newcommand{\CLASSINPUTinnersidemargin}{0.75in}
\newcommand{\CLASSINPUToutersidemargin}{0.75in}

\documentclass[letterpaper,10pt,conference]{IEEEtran}

\usepackage{cite}
\usepackage{amsmath,amssymb,amsfonts}
\usepackage{algorithmic}
\usepackage{graphicx}
\usepackage{textcomp}
\usepackage{xcolor}
\usepackage{booktabs}
\usepackage{multirow}
\usepackage{url}
\usepackage{balance}
\usepackage[hidelinks]{hyperref}
\usepackage{subcaption}

\begin{document}

\title{\vspace{0.25in}SURGE: Sonar-fUsed Reconstruction and localization via image-gated Graph Estimation}

\author{
\IEEEauthorblockN{
Mohammed Ibrahim M\IEEEauthorrefmark{1},
Vallabh Deogaonkar\IEEEauthorrefmark{2},
Trung Dong\IEEEauthorrefmark{1},
Jane Shin\IEEEauthorrefmark{3}
Abhilash Somayajula\IEEEauthorrefmark{2},
Xiaomin Lin\IEEEauthorrefmark{1}
}
\IEEEauthorblockA{
\IEEEauthorrefmark{1}
\textit{Department of Electrical and Computer Engineering}\\
\textit{University of South Florida}\\
Tampa, FL, USA\\
mohammedibrahimm@usf.edu
}
\IEEEauthorblockA{
\IEEEauthorrefmark{2}
\textit{Department of Ocean Engineering}\\
\textit{Indian Institute of Technology Madras}\\
Chennai, India\\
}
\IEEEauthorblockA{
\IEEEauthorrefmark{3}
\textit{Department of Naval Architecture and Ocean Engineering}\\
\textit{Seoul National University}\\
Seoul, South Korea\\
}
}

\maketitle


\begin{abstract}
Remotely operated vehicles (ROVs) are widely used to explore and inspect underwater environments such as caves, shipwrecks, and submerged infrastructure. These missions require accurate 3D understanding of the surrounding environment, which depends on both reliable vehicle localization and metric scene reconstruction. However, external positioning is often unavailable underwater, requiring small ROVs to rely primarily on onboard perception. Optic vision provides rich visual and geometric information but suffers from scale ambiguity and trajectory drift, whereas 2D imaging sonar provides metric range but incomplete 3D geometry. Existing underwater reconstruction approaches typically address these limitations separately or assume known sensor poses, leaving localization and reconstruction disconnected. We present \textbf{SURGE}, a camera sonar framework that jointly estimates the ROV trajectory and target location by integrating visual and acoustic observations within a factor graph, then uses the recovered metric poses for sonar Gaussian splatting. Experiments on real underwater RGB sonar observations show that \textbf{SURGE} substantially improves localization consistency over conventional vision based pose estimation and produces a more compact, natively metric reconstruction than RGB Gaussian splatting baselines.
\end{abstract}


\section{Introduction}

Remotely operated vehicles are increasingly used for underwater exploration and inspection of environments such as caves, shipwrecks, and submerged infrastructure~\cite{hou2022underwater,labra2023net,hurtos2017underwater,molina2024containers}. These missions require accurate three dimensional understanding of the environment for navigation, inspection, and spatial mapping. Building such representations, however, depends on both reliable vehicle localization and metric scene geometry.

\begin{figure}[t]
\centering
\includegraphics[width=0.99\columnwidth]{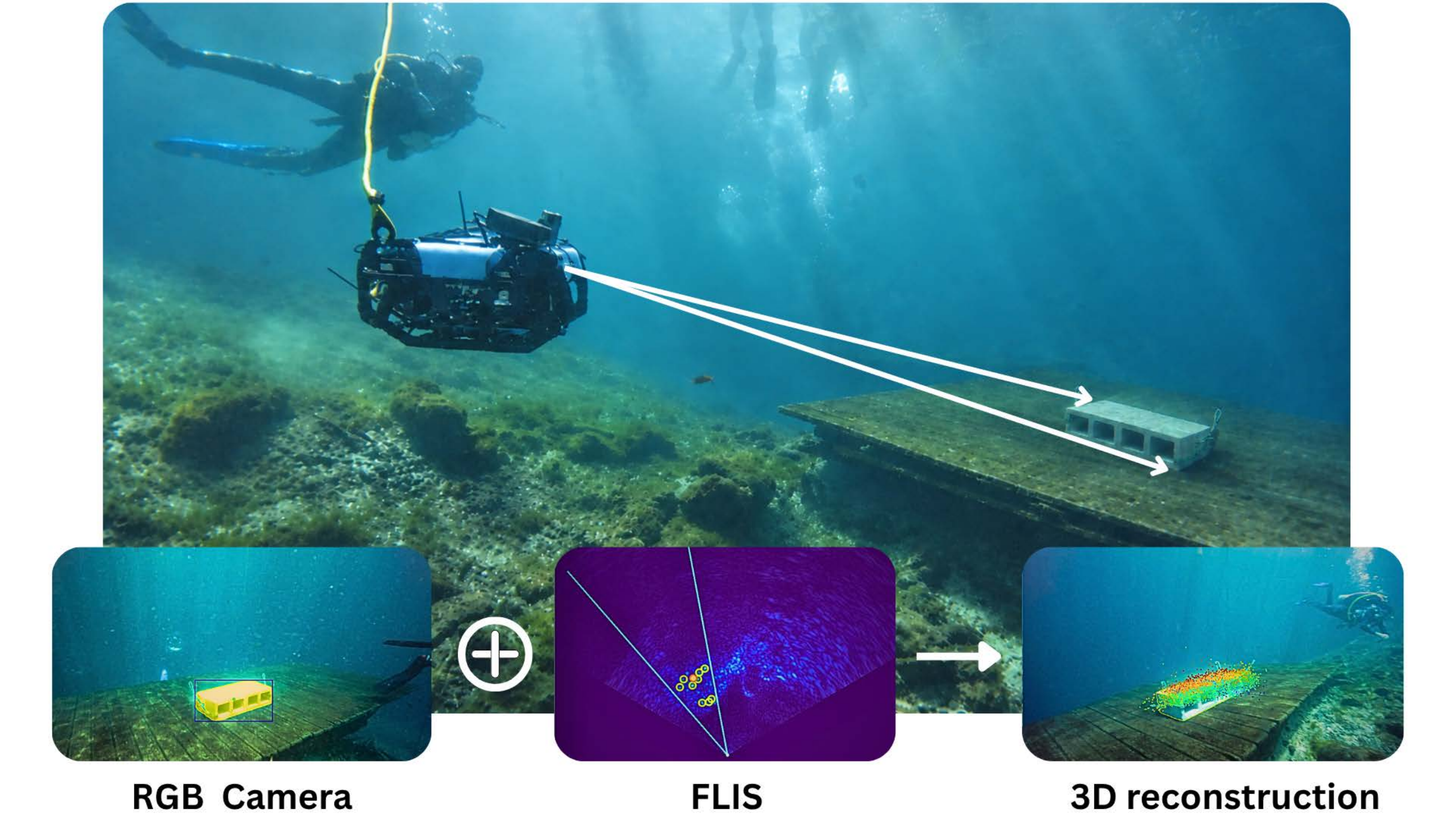}
\caption{A representative scenario of an ROV scanning a
cinder-block target, with the imaging sonar's and camera's
fields of view projected onto the object for localization
and reconstruction.}
\label{fig:teaser}
\end{figure}

Underwater localization is particularly challenging because external positioning is often unavailable, forcing small ROVs to rely on onboard perception~\cite{paull2014auvnav}. Monocular cameras provide rich visual and geometric information but suffer from scale ambiguity and trajectory drift, especially when visibility and feature quality degrade underwater~\cite{akkaynak2019sea}. Forward looking imaging sonar provides complementary metric range measurements and remains effective under poor visibility, but conventional two dimensional sonar observes range and azimuth while leaving elevation ambiguous~\cite{huang2015acousticsfm,negahdaripour2007optiacoustic}. Neither modality alone therefore provides both reliable metric localization and complete scene geometry, as shown in Fig. \ref{fig:teaser}.

Recent acoustic optical methods such as AONeuS and Z Splat demonstrate the benefit of combining camera and sonar measurements for underwater reconstruction~\cite{aoneus2024,zsplat2024}. However, these approaches largely assume known sensor poses, while other methods address pose estimation and reconstruction primarily within a single sensing modality~\cite{sonarsplat2025,seasplat2025,watersplatting2025}. This leaves a gap between multimodal reconstruction and the localization needed to support it in real underwater operation.

This separation is limiting in practice because reconstruction quality depends directly on the accuracy and metric consistency of the estimated sensor trajectory. We present \textbf{SURGE}, a camera sonar framework for joint underwater localization and metric reconstruction. Our contributions are threefold:

\begin{enumerate}
\item We introduce a multimodal factor graph that jointly estimates the ROV trajectory and the location of a compact underwater target from monocular RGB and two dimensional forward looking imaging sonar observations.

\item We develop a multi view sonar aggregation strategy and a target extent aware acoustic residual that provide stable metric constraints for joint pose and target estimation.

\item We use the recovered metric poses for target-scoped sonar
Gaussian splatting and demonstrate improved localization consistency
and more compact metric reconstruction than RGB Gaussian-splatting
baselines on real underwater data.
\end{enumerate}

\section{Related Work}
\label{sec}

\begin{table*}[t]
\centering
\caption{Comparison of representative underwater 3D reconstruction methods.
``Real'' denotes data collected outside controlled tank environments.}
\label{tab:landscape}

\small
\renewcommand{\arraystretch}{1.15}
\setlength{\tabcolsep}{9pt}

\begin{tabular}{@{}cccccc@{}}
\toprule
\textbf{System} &
\textbf{Modalities} &
\textbf{Poses} &
\textbf{Metric} &
\textbf{Joint Loc.+Recon.} &
\textbf{Setting} \\
\midrule

AONeuS~\cite{aoneus2024}
& RGB + Sonar
& GT
& Yes
& No
& Tank \\

Z Splat~\cite{zsplat2024}
& RGB + Sonar
& GT
& Yes
& No
& Tank \\

\addlinespace[2pt]

SonarSplat~\cite{sonarsplat2025}
& Sonar
& Est.
& Yes
& Yes
& Tank + River \\

Pose free 3D~\cite{posefreesonar2026}
& Sonar
& Est.
& Yes
& Yes
& Tank \\

\addlinespace[2pt]

SeaSplat~\cite{seasplat2025}
& RGB
& COLMAP
& No
& No
& Real \\

WaterSplatting~\cite{watersplatting2025}
& RGB
& COLMAP
& No
& No
& Real \\

\midrule

\textbf{SURGE}
& \textbf{RGB + FLIS}
& \textbf{Joint}
& \textbf{Yes}
& \textbf{Yes}
& \textbf{Real} \\

\bottomrule
\end{tabular}

\end{table*}

Recent underwater reconstruction methods differ primarily in two aspects: the sensing modalities they use and whether sensor poses are known or must be estimated from the observations. These distinctions are important for SURGE because accurate reconstruction depends on both complementary visual acoustic geometry and a reliable metric trajectory. Table~\ref{tab:landscape} summarizes representative approaches and the setting considered in this work.

\subsection{Multi-modal Reconstruction with Known Poses}

Camera and sonar provide complementary information for underwater reconstruction, motivating methods that combine optical appearance with acoustic geometry. AONeuS~\cite{aoneus2024} and Z-Splat~\cite{zsplat2024} demonstrate that acoustic measurements can improve three dimensional reconstruction when optical observations are degraded by underwater visibility. Collado-Gonzalez~\emph{et al.} similarly combine image regions with sonar range measurements for reconstruction in turbid environments~\cite{collado2025optiacoustic}. These methods establish the value of acoustic optical fusion, but primarily assume externally provided sensor poses. As a result, they address reconstruction once the trajectory is known rather than recovering localization and reconstruction from the same measurements.

\begin{figure*}[t]
\centering
\vspace{4mm}
\includegraphics[width=\textwidth, trim=0 2cm 0 2.6cm, clip]{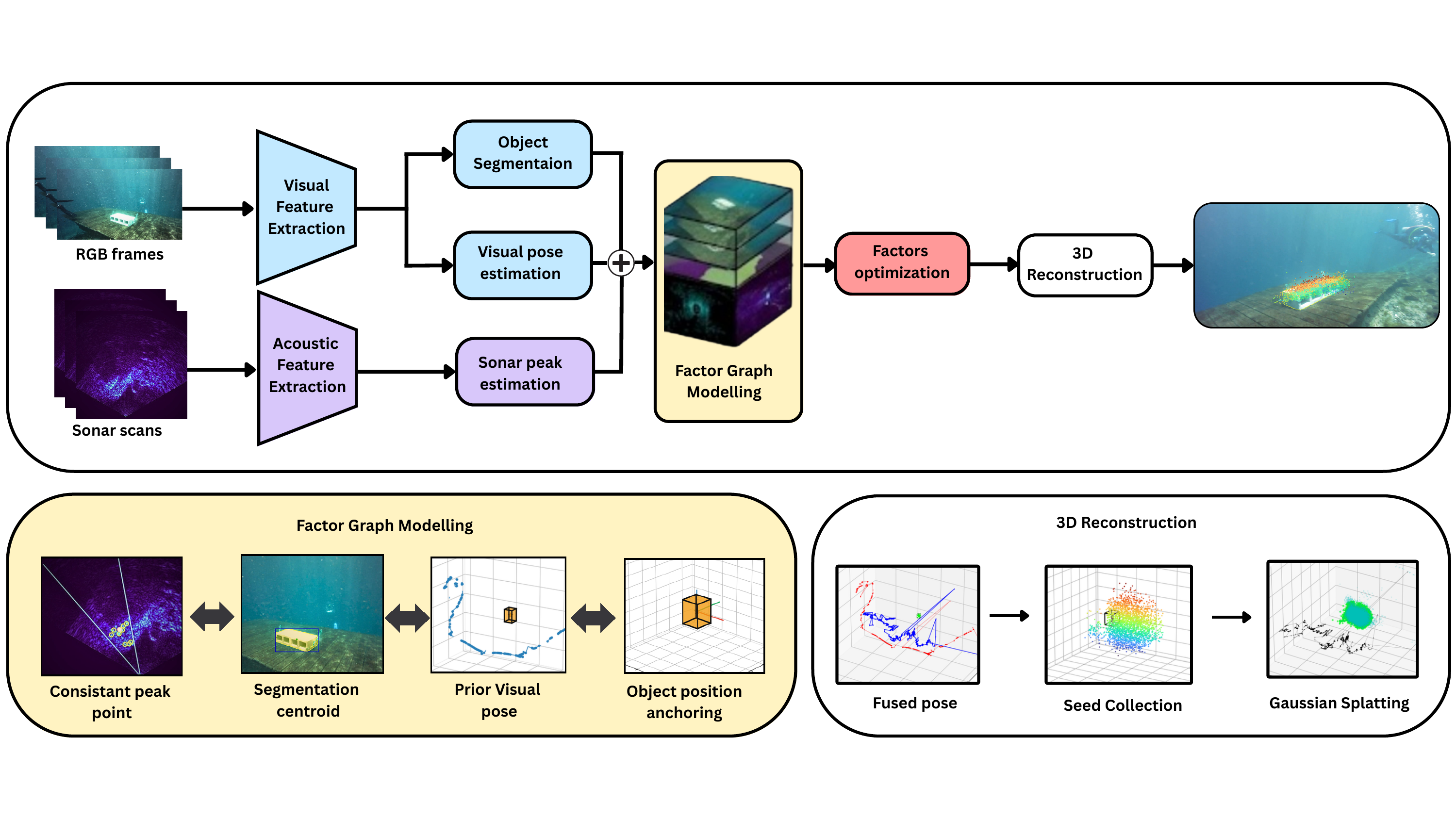}
\caption{Pipeline overview.
RGB frames and sonar scans pass through visual and acoustic
feature extraction into object segmentation, visual pose
estimation, and sonar peak estimation. Factor-graph modelling
fuses a multi-view consistent peak, the segmentation centroid,
a prior visual pose, and an object-position anchor; factor
optimization recovers metric ROV poses and the target landmark.
Cone-arch seed collection and 2D-sonar Gaussian splatting then
produce the reconstruction reprojected onto RGB.}
\label{fig:pipeline}
\end{figure*}

\begin{figure}[t]
\centering

\includegraphics[width=0.95\linewidth]{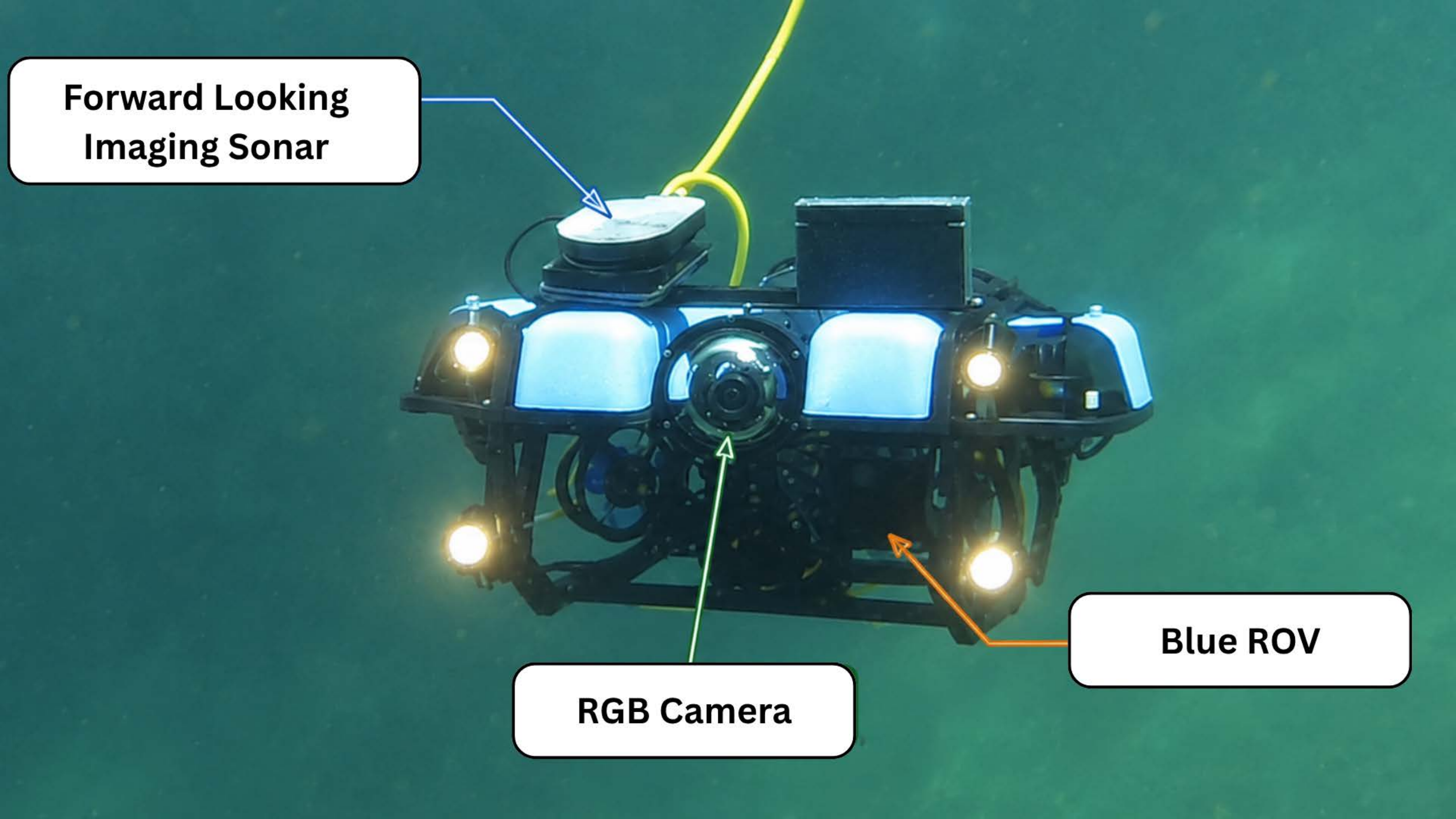}
\caption{The data collection platform. A BlueROV2 carries an RGB
camera and a 2D forward looking sonar in a rigid, forward facing configuration.}
\label{fig:rov}

\vspace{-4mm}
\end{figure}

\subsection{Single-Modality Reconstruction with Estimated Poses}

A parallel line of work relaxes the requirement for known poses, but generally within a single sensing modality. SonarSplat~\cite{sonarsplat2025} extends Gaussian splatting to forward-looking sonar while jointly optimizing the acoustic scene representation and sensor poses. Differentiable space carving reconstructs occupancy from imaging-sonar observations when poses are available~\cite{feng2024dsc}, while other approaches address reconstruction under pose drift~\cite{acoustic_drift_2025} or recover sonar geometry without supplied poses using priors from vision foundation models~\cite{posefreesonar2026}. Together, these methods show that acoustic observations can support both pose reasoning and reconstruction despite the geometric ambiguity of imaging sonar.

For optical sensing, SeaSplat~\cite{seasplat2025} and WaterSplatting~\cite{watersplatting2025} reconstruct real underwater scenes from RGB imagery using COLMAP poses, while AquaSplat~\cite{aquasplat2026} introduces a dual-branch representation for underwater RGB reconstruction. These methods improve the representation of underwater appearance and geometry, but their reconstruction remains tied to trajectories recovered from visual Structure-from-Motion. Thus, pose-aware reconstruction has been explored for both optical and acoustic sensing, but primarily without using their complementary measurements jointly to recover a metric trajectory.

\subsection{Underwater Factor-Graph SLAM}

Underwater SLAM provides the complementary perspective of estimating vehicle motion from multiple sensor constraints. Existing systems combine acoustic range, inertial sensing, and optical observations to estimate trajectories and map features~\cite{ferreira2026slam,mcconnell2022perception,li2018flsposegraph,westman2018sonarslam}. Imaging-sonar multi-robot systems further recover trajectories and object-level maps from multibeam measurements~\cite{huang2025dracoslam2,mcconnell2022dracoslam}. These approaches demonstrate the value of acoustic constraints for underwater localization, but are not designed to connect monocular RGB and 2D FLIS localization directly to downstream target reconstruction.

The observation model is also different for compact extended targets. Conventional landmark formulations typically associate acoustic observations with discrete features, whereas the dominant return from an extended object can move across its visible surface as the viewpoint changes. SURGE therefore models the sonar observation relative to the spatial extent of the target rather than forcing every acoustic return to coincide with a single geometric center.

\subsection{Position of the Present Work}

Prior work therefore establishes three complementary capabilities: acoustic optical fusion can improve underwater reconstruction when poses are known, single-modality methods can reconstruct scenes while estimating or tolerating pose uncertainty, and underwater SLAM can recover vehicle trajectories from acoustic and visual constraints. What remains less explored is their combination: jointly estimating metric vehicle motion and target location from monocular RGB and 2D forward-looking sonar, then using the recovered trajectory for reconstruction in real underwater environments without external positioning.

SURGE addresses this gap by coupling visual and acoustic observations within a factor graph and using the resulting metric poses for target-scoped sonar Gaussian splatting. The framework therefore connects localization and reconstruction through the same complementary camera-sonar measurements rather than treating accurate poses as an independent prerequisite.

\section{Method}
\label{sec:method}
At a high level, SURGE recovers a metric ROV trajectory and a
static target centre from paired RGB and 2D sonar frames, then
reconstructs the target with sonar Gaussian splatting under those
poses. Vision tells \emph{where} the object appears in the image;
sonar tells \emph{how far} it is. The factor graph fuses both,
and the reconstruction stage consumes only the resulting poses.
The pipeline is summarised in Fig.~\ref{fig:pipeline}.
Segmentation and SfM are treated as fixed inputs; the method
itself has four stages aligned with that figure.
Sec.~\ref{sec:rig} states the sensing configuration, lever arm,
and assumed target half-extents~$h$.
Sec.~\ref{sec:orange} builds the acoustic observation: RGB-gated
peak candidates on the FLIS fan are aggregated by a multi-view
vote into one world-consistent orange flash, which is then
reprojected into each frame, matching the
feature-extraction and sonar peak estimation path in
Fig.~\ref{fig:pipeline}.
Sec.~\ref{sec:graph} jointly estimates metric camera poses against
that flash, a frozen landmark box of extents~$h$, and RGB
centroid/box factors inside the
factor-graph modelling / factors-optimization block.
An important modelling choice in that graph is the polar disk
sonar residual: because the dominant FLIS return typically sits on
a near face rather than at the geometric centre~$b$, a hard
centre-to-peak residual is inconsistent, and the disk tolerates
that offset without discarding the measurement.
Sec.~\ref{sec:recon} then runs cone-arch seed collection and
2D-sonar Gaussian splatting on the recovered poses
(Fig.~\ref{fig:pipeline}).

\subsection{Sensors, frames, and inputs}
\label{sec:rig}

Data are collected with a BlueROV2 carrying a forward-facing RGB
camera and a 2D forward-looking imaging sonar in a rigid
configuration (Fig.~\ref{fig:rov}). The camera--sonar extrinsics
are treated as known and fixed. COLMAP provides an initial,
soft visual pose prior; SAM~\cite{sam2023} provides a per-frame
target mask, bounding box, and centroid. We represent the target
as a static upright box of assumed half-extents
$h{=}[0.10,0.10,0.20]$\,m; the centre~$b$ is unknown and estimated.
Only yaw is refined on top of COLMAP; pitch and roll follow SfM.

\subsection{Sonar measurement layer}
\label{sec:orange}
\begin{figure}[t]
\centering
\vspace{4mm}
\includegraphics[width=\linewidth]{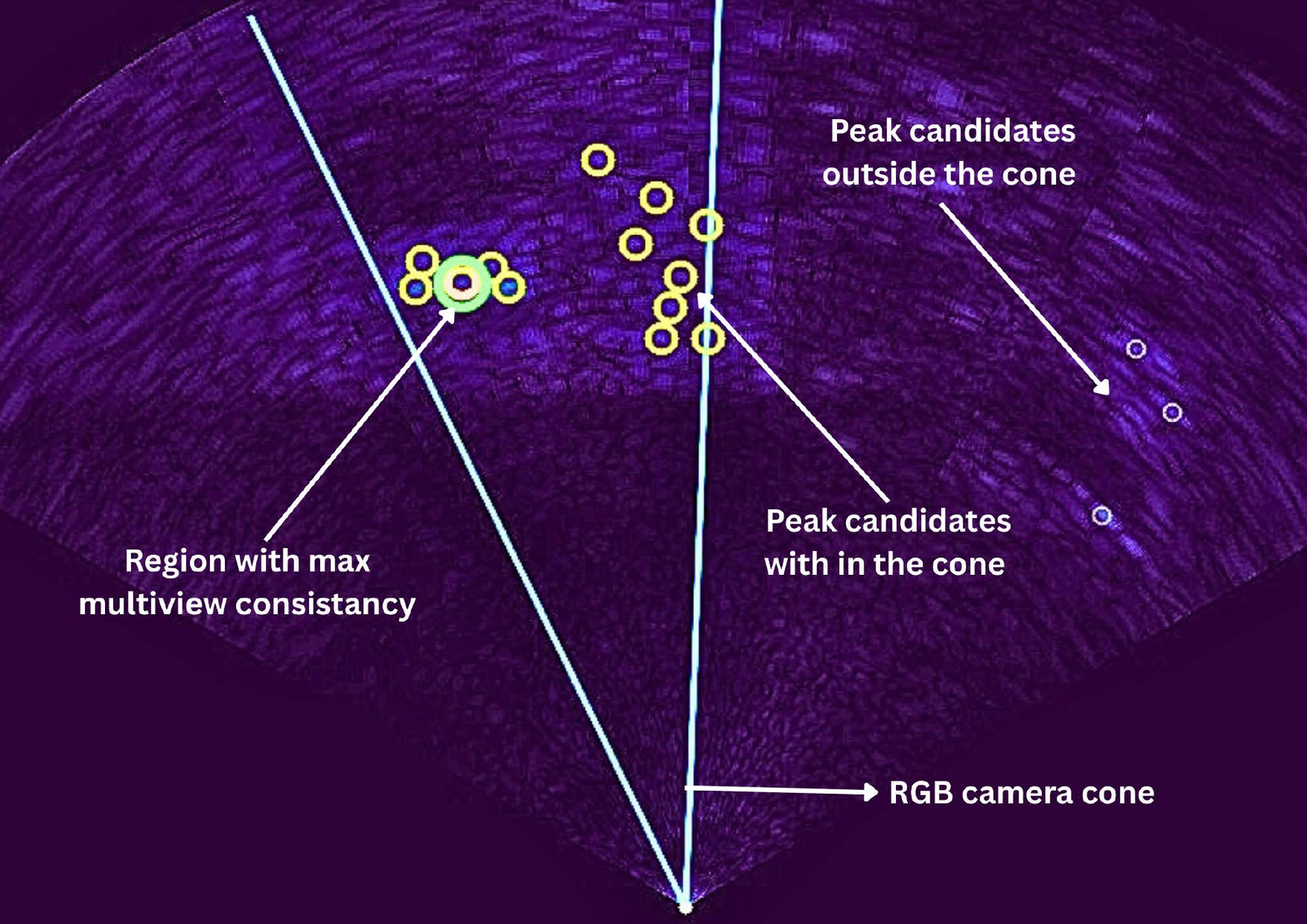}
\caption{Sonar fan with candidate peaks. The RGB camera cone
constrains admissible peaks to the target's angular extent; peaks
outside the cone are discarded. Among in-cone candidates, the
multi-view vote elects the frame's sonar observation.}
\label{fig:peaks}
\end{figure}

A brightest-bin factor fails under clutter (rock, tether,
sidelobe). We build one acoustic observation per frame in three
steps (Fig.~\ref{fig:peaks}); the same flow appears in the
feature-extraction and peak-estimation stages of
Fig.~\ref{fig:pipeline}. Raw sonar peaks are noisy: clutter, tether, and sidelobes can
outshine the target. We therefore build one \emph{consensus}
acoustic observation per frame--a range--azimuth measurement
that is consistent across views. In figures this locked return is
drawn in orange, so we refer to it as the orange observation; it
is \emph{not} a second physical sensor.

\emph{Step 1 --- local peaks.} Detect local maxima with
non-maximum suppression inside a range gate.

\emph{Step 2 --- RGB cone gate.} Project the SAM mask through the
lever into polar $(r,\mathrm{az})$ and discard peaks outside that
cone.

\emph{Step 3 --- multi-view consensus.} In-cone peaks still walk
over faces. For pairs of frames $(i,j)$, intersect frame~$i$'s
sonar elevation arc with frame~$j$'s RGB-centroid ray, bin the
hits in $18$\,cm voxels, and take the mode as a locked world
point~$o^\star$. Reprojecting $o^\star$ into each fan gives that
frame's orange observation: world-consistent, but still
pose-dependent. Of $881$ cone-gated frames, $880$ retain a
consensus flash.

\subsection{Factor graph}
\label{sec:graph}

\begin{figure}[t]
\centering
\includegraphics[width=\linewidth]{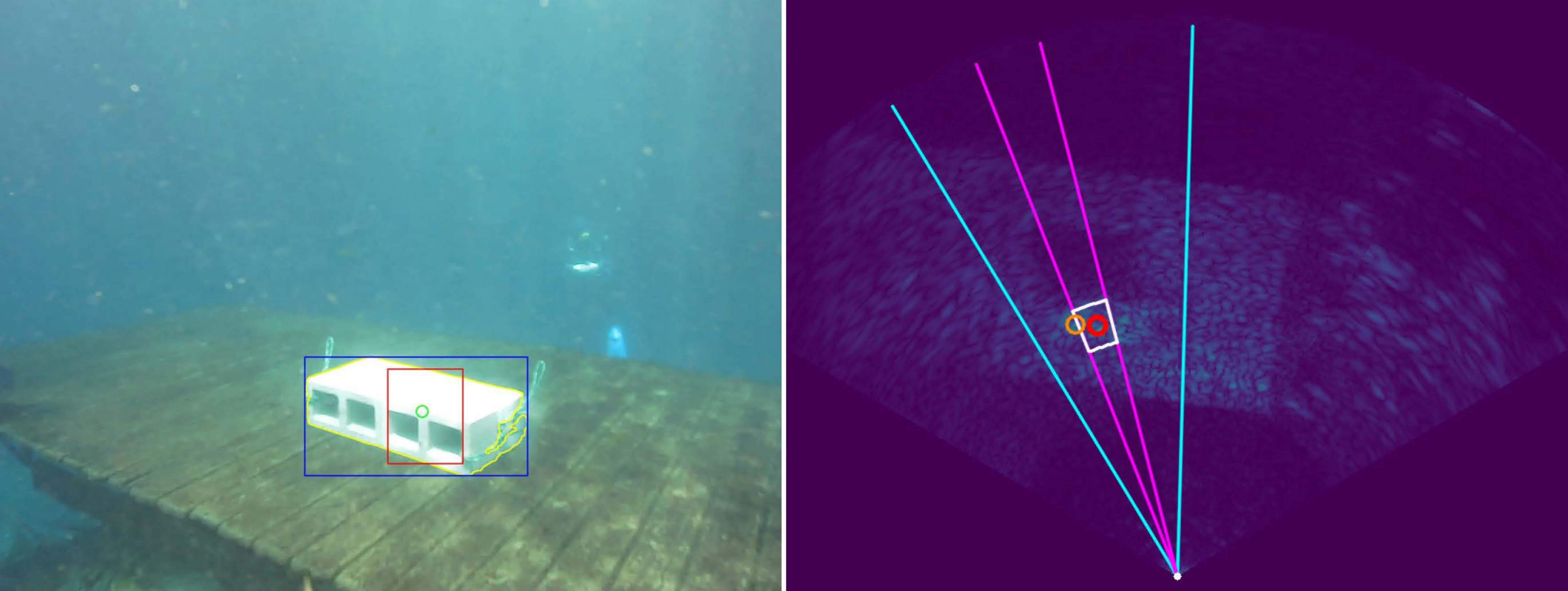}
\caption{Correspondence between RGB and sonar observations for
a representative frame. Left: RGB with SAM silhouette (yellow),
mask AABB (blue), and mask centroid (green). Right: matching FLIS
fan with RGB-mask cone (cyan), factor-graph cone (magenta),
projected brick (white), locked orange return, and projected
centre~$b$ (red).}
\label{fig:modalities}
\end{figure}

The factor graph estimates the camera centres~$\{c_k\}$, small
yaw corrections~$\{\delta\psi_k\}$, and one static target
centre~$b$, so that RGB and sonar observations agree on the same
metric object. We jointly estimate per-frame centres~$c_k$, yaw corrections
$\delta\psi_k$ ($R_k=R_z(\delta\psi_k)\,R_k^{\mathrm{C}}$), and one
static landmark centre~$b$. Assumed half-extents~$h$ are held
fixed. This is cast as a factor-graph
estimation~\cite{dellaert2017factorgraphs}. Pitch and roll follow COLMAP. The first valid frame is
gauge-fixed to its COLMAP centre with $\delta\psi_0=0$. The full
factor set is listed in Table~\ref{tab:factors}; residuals other
than the elevation gate are Huber-robustified.
Fig.~\ref{fig:modalities} shows the RGB and sonar observations
that feed $F_{\mathrm{bbox}}$, $F_{\mathrm{uv}}$, and
$F_{\mathrm{sonar}}$; the remaining factors are pose priors and
geometric gates and are not drawn separately. In Fig.~\ref{fig:pipeline}, this corresponds to fusing the
consistent peak, segmentation centroid, prior visual pose, and
object-position anchor, then optimizing the factors.

\begin{table}[t]
\centering
\vspace{4mm}
\caption{Factors in the reported configuration.}
\label{tab:factors}
\scriptsize
\begin{tabular}{@{}lcp{5.8cm}@{}}
\toprule
Factor & Weight & Residual\\
\midrule
$F_{\mathrm{sonar}}$ & $4.0$
  & Projected $b$ vs.\ orange; zero cost inside a $12$\,cm
    polar disk.\\
$F_{\mathrm{bbox}}$ & $1.8$
  & Projected AABB of $b\pm h$ vs.\ SAM box.\\
$F_{\mathrm{uv}}$ & $1.5$
  & Projected $b$ vs.\ SAM mask centroid.\\
$F_{\mathrm{colmap}}$ & $0.25$
  & Soft absolute anchor to COLMAP ($\sigma_c{=}0.8$\,m);
    down-weighted near teleports.\\
$F_{\mathrm{smooth}}$ & $2.5$
  & Small $\Delta c$; matches $\Delta c^{\mathrm{C}}$ when
    $\|\Delta c^{\mathrm{C}}\|\le 0.4$\,m.\\
$F_{\mathrm{yaw}}$ & $0.7$
  & Small $\delta\psi$ and $\Delta\delta\psi$.\\
$F_{\mathrm{elev}}$ & hard
  & $|\mathrm{el}(b)|\le 25^\circ$.\\
$F_{\mathrm{depth}}$ & $0.05$
  & $b$ in front of the camera.\\
\bottomrule
\end{tabular}
\end{table}

\paragraph{Disk sonar factor ($F_{\mathrm{sonar}}$).}
The dominant sonar return usually comes from a near face of the
block, not from its geometric centre. A residual that forces the
peak to equal~$b$ is therefore inconsistent; we instead allow a
small polar tolerance (the disk) around the orange observation. Let $p(r,\theta)=(r\cos\theta,\,r\sin\theta)$ map polar
quantities to the fan plane, $(r_k^b,\theta_k^b)$ the projection
of~$b$, and $(r_k^\circ,\theta_k^\circ)$ the locked orange. With
disk radius $d{=}12$\,cm (on the order of the assumed box
half-extents),
\begin{equation}
e_k^{\mathrm{sonar}}
=
\frac{1}{\sigma_r}\,
\mathrm{ReLU}\!\Bigl(
\bigl\|p(r_k^b,\theta_k^b)-p(r_k^\circ,\theta_k^\circ)\bigr\|_2
- d\Bigr).
\label{eq:disk}
\end{equation}
Cost is zero while the projected centre lies inside the disk about
the flash; outside, a Huber pull acts on the cameras. This lets a
face return disagree with~$b$ without dropping the observation.
The factor is always active.

\paragraph{RGB factors ($F_{\mathrm{bbox}}$, $F_{\mathrm{uv}}$).}
$F_{\mathrm{bbox}}$ compares the projected AABB of $b\pm h$ to the
SAM box (four residuals per frame). $F_{\mathrm{uv}}$ compares the
projected centre to the mask centroid. Together they constrain
bearing without overpowering the metric sonar disk
(Fig.~\ref{fig:modalities}).

\paragraph{COLMAP prior ($F_{\mathrm{colmap}}$).}
A weak absolute anchor ($\sigma_c{=}0.8$\,m) keeps the trajectory
near the SfM shape. When COLMAP-reported inter-frame motion
exceeds $0.9$\,m, nearby weights are scaled by $0.05$ so teleports
are not copied.

\paragraph{Smoothness ($F_{\mathrm{smooth}}$).}
Prefers small $\Delta c$ between consecutive frames and, when
COLMAP motion is below $0.4$\,m, also matches
$\Delta c^{\mathrm{C}}$.

\paragraph{Yaw ($F_{\mathrm{yaw}}$).}
Penalises large $\delta\psi_k$ and large frame-to-frame
$\Delta\delta\psi$, limiting heading drift that would otherwise
misplace the orange laterally at range.

\paragraph{Elevation and depth ($F_{\mathrm{elev}}$, $F_{\mathrm{depth}}$).}
$F_{\mathrm{elev}}$ is a hard gate keeping $b$ inside the FLIS
aperture ($|\mathrm{el}(b)|\le 25^\circ$). $F_{\mathrm{depth}}$ is a
weak prior that $b$ remains in front of the camera.

We freeze~$b$ after a short box-only warm start so cameras move
around a static object. Optimisation is GPU Adam, $400$
iterations, learning rate $10^{-3}$. All factors remain active;
there is no staged schedule.

\subsection{Reconstruction}
\label{sec:recon}

The polar renderer follows~\cite{oceans2026fssplat}; we do not claim
a new splatting algorithm. SURGE contributes estimated metric poses
and target-scoped cone-arch seeding on real data. Given metric poses and~$b$, we reconstruct only the target
neighbourhood with 2D-sonar Gaussian splatting. Seeding is
restricted to short elevation arches near~$b$ (cone-arch seeds)
so that the cloud stays on the object rather than filling the
seafloor. With metric poses and landmark $b$ recovered (fused pose in
Fig.~\ref{fig:pipeline}), we reconstruct the target by 2D
forward-looking imaging sonar Gaussian splatting
~\cite{oceans2026fssplat} via seed collection and Gaussian
splatting as shown in that figure.

\paragraph{Elevation ambiguity.}
An FLIS ping reports intensity $I(r,\theta)$ over range and azimuth
only. A strong return at $(r,\theta)$ is consistent with any point on
the elevation arc inside the vertical beam,
\begin{equation}
\mathcal{A}(r,\theta)
=
\bigl\{\mathbf{x}:
\|\mathbf{x}\|=r,\;
\mathrm{az}(\mathbf{x})=\theta,\;
\phi\in[\phi_{\min},\phi_{\max}]\bigr\},
\label{eq:arc}
\end{equation}
with $[\phi_{\min},\phi_{\max}]=[-25^\circ,25^\circ]$ on the M750d.
Reconstruction therefore cannot treat a bin as a 3D point; mass must
be hypothesised along~$\phi$ and resolved by multi-view consistency.

\paragraph{Scene representation and polar rendering.}
The scene is a set of anisotropic Gaussians
\begin{equation}
\mathcal{G}=\{\mathcal{G}_i\}_{i=1}^{N},\qquad
\mathcal{G}_i=(\boldsymbol{\mu}_i,\boldsymbol{\Sigma}_i,\alpha_i,\rho_i),
\label{eq:gaussians}
\end{equation}
where $\boldsymbol{\mu}_i$ is the mean, $\boldsymbol{\Sigma}_i$ the
covariance, $\alpha_i$ opacity, and $\rho_i$ a scalar acoustic
reflectivity (no RGB appearance). Following~\cite{oceans2026fssplat},
each mean is transformed into the sonar frame, linearised into
$(r,\theta,\phi)$ with the spherical Jacobian used
in~\cite{sonarsplat2025}, reduced to the polar $(r,\theta)$ plane, and
$\alpha$-composited along range to form a rendered fan
$\hat{I}_k(r,\theta)$. Training minimises
\begin{equation}
\mathcal{L}
=
\sum_k
\bigl\|\hat{I}_k - I_k\bigr\|_1
+
\lambda_\alpha\,\mathcal{R}_\alpha
+
\lambda_s\,\mathcal{R}_s
+
\lambda_\phi\,\mathcal{R}_\phi,
\label{eq:gs-loss}
\end{equation}
where $\mathcal{R}_\alpha$, $\mathcal{R}_s$, and $\mathcal{R}_\phi$
are the opacity, scale, and elevation-ray regularisers from the 2d Sonar Splatting \cite{oceans2026fssplat}
pipeline (discouraging empty mass and thin vertical tubes along a
single $(r,\theta)$).

\paragraph{What we reuse from~\cite{oceans2026fssplat}.}
Elevation-arc initialisation, the polar renderer, and three always-on
components: elevation gating in the visibility mask, a creation-round
replay filter on the per-Gaussian information matrix, and
$\lambda_{\min}$ scoring for opacity weighting. Those elements were
validated in simulation with known poses (fence / truss); we keep the
same rendering backbone here.

\paragraph{What changes in this work.}
The 2d Sonar Splatting \cite{oceans2026fssplat} setting assumes known sensor poses and can seed across the
observed scene. Here the cameras are the factor-graph estimate of
Sec.~\ref{sec:graph}, not ground truth, and poses are held fixed
during splatting (we do not jointly refine pose inside the renderer).
Seeding is restricted to short elevation arches on cone-gated peaks
near the localised box $b\pm h$ (plus a small margin), with initial
covariance scale $6$\,cm (about one range bin). Full-fan seeding
produces a seafloor pancake; seeds smaller than one bin receive no
usable polar gradient. Training uses unmasked
$128{\times}128$ polar fans for $15$ epochs. The output is therefore
a compact, metric target cloud under real freshwater observations,
conditioned on the joint pose--landmark solution rather than on
simulated ground-truth pose.
\section{Experiments}
\label{sec:exp}

\subsection{Dataset and protocol}
\label{sec:data}

We record $1{,}199$ RGB--FLIS pairs at a freshwater spring-cave
site in the southeastern United States, of one cinder block on a
submerged deck, with the vehicle orbiting at $2$--$5$\,m range at
approximately similar depth. Clear freshwater
means reported failure modes reflect the estimator rather than
turbidity. Camera calibration used by COLMAP and the graph
follows Sec.~\ref{sec:rig}. After lockstep
synchronisation and dropping a large COLMAP teleport window
(sids $967$--$969$), $880$ frames remain; $874$ enter the graph as
inliers. The pipeline is offline: every valid frame is a training
view for both the graph and the reconstruction. Assumed
half-extents follow Sec.~\ref{sec:rig}; we do not use a surveyed
brick size as ground truth.

\subsection{Baselines and metrics}
\label{sec:base}
We evaluate localization of the joint graph and compactness of
the sonar reconstruction near~$b$. We do not claim a
face-accurate mesh or a full-scene map.
We compare against RGB Gaussian-splatting baselines trained on
COLMAP poses:

\emph{Vanilla 3DGS}~\cite{kerbl3Dgaussians}: no underwater medium
model.

\emph{SeaSplat}~\cite{seasplat2025}: underwater medium model;
same inputs.

\emph{WaterSplatting}~\cite{watersplatting2025}: alternative
underwater-medium formulation; same inputs.

\emph{AquaSplat}~\cite{aquasplat2026}: additional underwater RGB
baseline under the same protocol. 

OASIS~\cite{phung2025oasis} is omitted: it assumes imaging sonar
and a known vehicle/arm pose, whereas this bag is 2D FLS without
external positioning. RGB clouds are not natively metric. Before scoring distance to
$b$, we similarity-align each RGB Gaussian cloud to the
factor-graph cameras with a free Sim(3) (Umeyama on camera
centres; recovered scale $\approx 0.38$). Our cone-arch cloud is
already in that metric frame and is not re-aligned.
For localization we report inlier count, median radial and azimuth
residuals of projected~$b$ against the consensus orange, and mean
projected-box vs.\ SAM-box error. For reconstruction we report the
median distance of opaque Gaussian means to~$b$, the fraction of
opaque means within $0.5$\,m of~$b$, mean projected mass inside
the SAM mask, and whether the cloud is natively metric.

\subsection{Localization}
\label{sec:loc}

Freezing COLMAP and moving only~$b$ fails (metre-scale range
error). Optimising poses with~$b$ fixed recovers most of the
gain; the centre$+$disk residual further reduces face bias and
box error. Table~\ref{tab:loc} reports graph configurations on the
$880$-frame sequence. The reported centre$+$disk formulation
retains $874/880$ frames at $10.6$\,cm median radial and
$3.33^\circ$ median azimuth residuals, with mean projected-box vs.\
SAM-box error of $93$\,px (on the order of SAM mask jitter). Because no surveyed vehicle trajectory is available, we report
consistency of the estimated poses with the multi-view-consistent
sonar observation and RGB segmentation, not absolute trajectory error.

\begin{table}[t]
\centering
\vspace{4mm}
\caption{Localization on the $880$-frame sequence.
$|\Delta r|$ and $|\Delta\mathrm{az}|$ are medians of projected
$b$ against locked orange; bbox is mean projected-box vs.\ SAM
error.}
\label{tab:loc}
\scriptsize
\setlength{\tabcolsep}{2.5pt}
\begin{tabular*}{\columnwidth}{@{\extracolsep{\fill}}lcccc@{}}
\toprule
Config & Inl. & $|\Delta r|$ & $|\Delta\mathrm{az}|$ & bbox\\
 & & (cm) & ($^\circ$) & (px)\\
\midrule
COLMAP frozen
  & $312$ & $189$ & $12.4$ & $210$ \\
Pose opt., $b$ fix
  & $872$ & $12.1$ & $3.60$ & $108$ \\
\textbf{Centre$+$disk}
  & $\mathbf{874}$
  & $\mathbf{10.6}$
  & $\mathbf{3.33}$
  & $\mathbf{93}$ \\
\bottomrule
\end{tabular*}
\end{table}

Freezing COLMAP poses and optimising only~$b$ locks scale error in:
median $|\Delta r|$ is $1.89$\,m. The adjustment must therefore move
the cameras. Optimising poses with~$b$ frozen already recovers most
of the gain; the disk residual further reduces face-bias and bbox
error. Both pose-optimising rows keep the RGB centroid and bbox
factors ($w_{\mathrm{uv}}{=}1.5$, $w_{\mathrm{bbox}}{=}1.8$); the
disk changes the sonar residual geometry, not the RGB factors.

\begin{figure}[t]
\centering
\includegraphics[width=0.98\columnwidth]{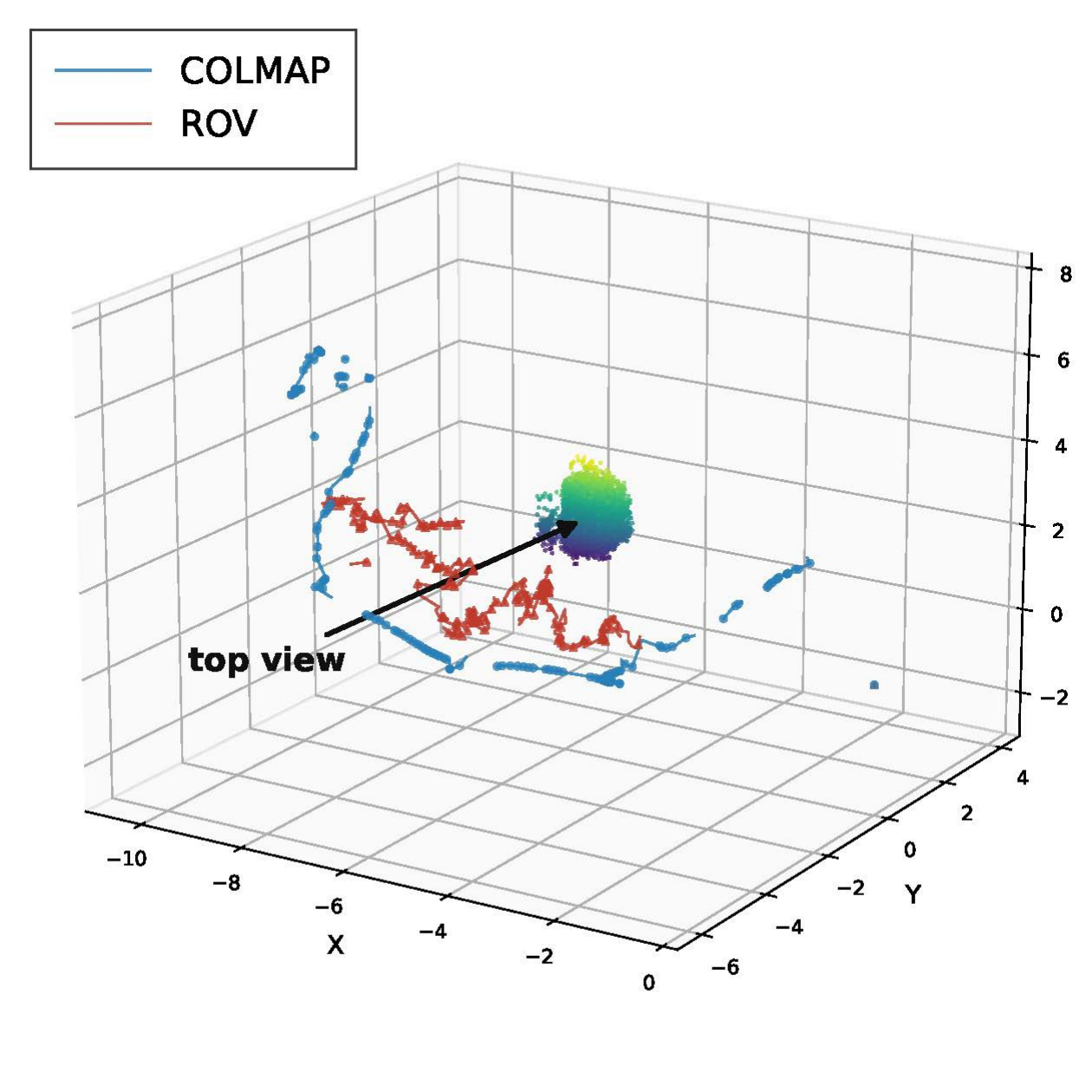}
\caption{Recovered ROV trajectory (red) and target landmark
(green) in the world frame, overlaid on the COLMAP prior (blue).
The factor graph metricises the COLMAP path against the consensus
orange and the frozen landmark, absorbing large SfM teleports
without copying them.}
\label{fig:traj}
\end{figure}

Fig.~\ref{fig:traj} shows COLMAP versus the recovered trajectory.
A $6.26$\,m COLMAP teleport (sids~$479$--$481$) appears as a
$1.70$\,m step: the graph accepts that motion occurred but does not
copy the jump. Early-sequence free frames can still move to place
the orange inside the disk under the gauge pin; that is expected
with a frozen landmark and an SfM-inconsistent initial pose.

\subsection{Quantitative reconstruction}
\label{sec:recon-exp}

RGB baselines are denser in image space but not metric; we
Sim(3)-align them to the factor-graph cameras before scoring
distance to~$b$. Our cloud is already in that frame. Table~\ref{tab:recon} compares reconstructions on the same $880$
frames. RGB baselines use frozen COLMAP poses; before scoring
distance to $b$ we similarity-align each RGB cloud to the
factor-graph cameras with Sim(3) (Umeyama on camera centres).
Even after alignment, opaque RGB means remain $1.4$--$2.2$\,m
from $b$, with at most $7.1\%$ of mass within $0.5$\,m and
$8$--$20\%$ mean in-mask projection. Our cone-arch splat is
natively metric in the same frame: median distance $0.65$\,m,
$30\%$ of opaque means within $0.5$\,m of $b$, and $36\%$ mean
in-mask projection. It is intentionally target-scoped, not a
full-scene map.

\begin{table}[t]
\centering
\caption{Reconstruction on the $880$-frame sequence (opaque
$\alpha\!\ge\!0.05$). RGB methods use COLMAP poses and are
Sim(3)-aligned to FG cameras before scoring; ours uses FG poses
and is natively metric. med$\to b$: median distance to landmark
$b$; $\le\!0.5$\,m: fraction within $0.5$\,m of $b$; In mask:
mean projected mass inside the SAM mask.}
\label{tab:recon}
\scriptsize
\setlength{\tabcolsep}{2.5pt}
\begin{tabular*}{\columnwidth}{@{\extracolsep{\fill}}lcccc@{}}
\toprule
Method & med$\to b$ (m) & $\le\!0.5$\,m & In mask & Metric\\
\midrule
Vanilla 3DGS   & $1.41$ & $6.8\%$ & $20\%$ & no\\
SeaSplat       & $2.20$ & $1.6\%$ & $8\%$  & no\\
WaterSplatting & $1.43$ & $7.1\%$ & $17\%$ & no\\
AquaSplat      & $1.43$ & $6.4\%$ & $17\%$ & no\\
\textbf{Ours}  & $\mathbf{0.65}$ & $\mathbf{30\%}$
 & $\mathbf{36\%}$ & \textbf{yes}\\
\bottomrule
\end{tabular*}
\end{table}

\begin{figure*}[t]
\centering
\vspace{4mm}
\includegraphics[width=\linewidth]{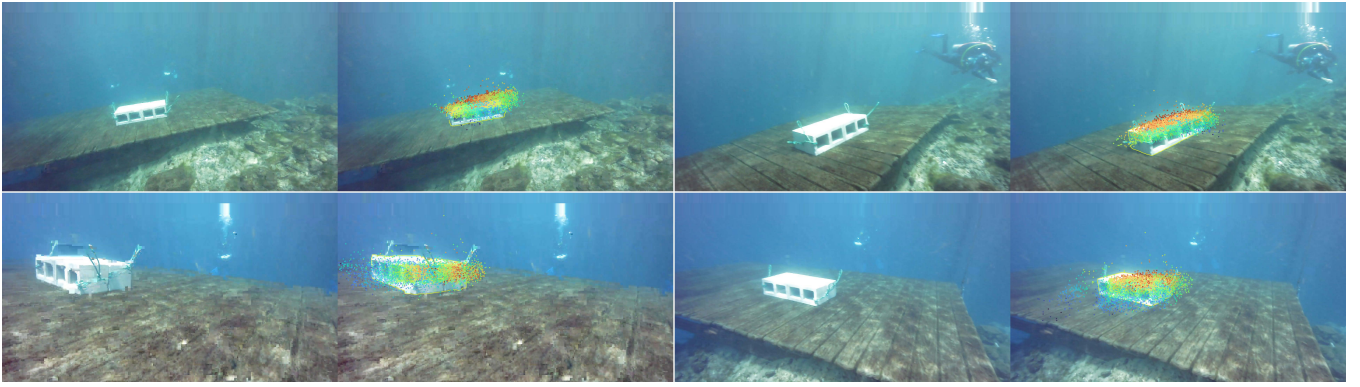}
\caption{Reconstruction projected into representative frames.
In each pair, raw RGB is left and the same view with projected
Gaussian centres is right. The cloud is compact and correctly
located near the target. Vertical spread follows the unmeasured
$\pm 25^\circ$ FLIS aperture under a mostly horizontal orbit.}
\label{fig:results}
\end{figure*}

Fig.~\ref{fig:results} shows overlays on representative viewpoints.
The cloud is not a face-accurate brick: vertical structure is
limited by sensor elevation ambiguity and the orbit, not by the
factor graph.

\section{Discussion}
\label{sec:discuss}

\subsection{What the pipeline recovers}
\label{sec:discuss-recover}

SURGE recovers metric-scale ROV poses, a static target centre, and
a compact sonar Gaussian cloud from a monocular camera and 2D FLIS
only (Fig.~\ref{fig:pipeline}). Median orange residuals are on the
order of one range bin, and the reconstruction is natively metric
and much closer to the localised landmark than RGB splatting on
COLMAP poses. The cloud is intentionally target-scoped: it is not a
face-accurate brick and not a full-scene map. Vertical structure is
limited by FLIS angular resolution, the unmeasured
$\pm 25^\circ$ elevation aperture, and a mostly horizontal orbit.

\subsection{Why the disk residual matters}
\label{sec:discuss-disk}

A hard centre-to-peak sonar residual is inconsistent for an extended
target: the dominant return usually sits on a near face, not at~$b$.
An earlier nearest-face model with Huber weighting looked correct but
silently down-weighted informative frames and collapsed range by about
a metre. Zeroing cost inside a polar disk of radius $d{=}12$\,cm
(on the order of the assumed half-extents) keeps those observations
active. When a measurement is useful but ambiguous within a modelling
tolerance, that tolerance belongs in the residual geometry, not only
in a robust weight.

\subsection{Limits of open-water reconstruction}
\label{sec:discuss-limits}

Denser or sharper geometry is hard to justify from this bag. The
orbit has little elevation diversity; FLIS provides no elevation
measurement; and the target occupies only a small fraction of each
fan. RGB can look detailed in image space yet miss metric scale,
while sonar alone leaves a vertical slab. Under these views and
sensors, a compact, correctly located, natively metric cloud near
the localised landmark is the strongest reconstruction claim we can
support.

\subsection{Discarded design alternatives}
\label{sec:not}

Several alternatives failed for clear reasons.
\emph{Peak-as-centre} cannot go to zero as the return walks over
faces, so~$b$ chases glints.
\emph{Freeze COLMAP, optimise only~$b$} locks scale error
(median $|\Delta r|{\approx}1.89$\,m).
\emph{Strong absolute COLMAP} copies multi-metre teleports.
\emph{Full-fan seeding} produces a seafloor pancake; seeds must stay
near~$b$.
\emph{Sub-bin initial covariance} removes the polar gradient and
collapses opacity within a few epochs.

\subsection{Scope and future work}
\label{sec:scope}
\label{sec:future}

Assumed half-extents~$h$ set both the box prior and the disk scale;
unknown-size targets would need a jointly estimated extent, which we
do not evaluate. The orange consensus assumes one compact static
object. Within small-ROV inspection of such a target with camera and
2D FLIS only, this is the setting we claim.

The main remaining limit is viewpoint diversity, not the renderer.
Closing the loop with active next-best-view control---motions that
reduce uncertainty in the joint pose--landmark estimate and in the
worst-constrained directions of the per-Gaussian information
matrix---is the natural continuation~\cite{oceans2026fssplat,ut2025uncertainty,wang2022virtualmaps}.

\section{Conclusion}
\label{sec:concl}

We presented SURGE, a camera--sonar factor-graph pipeline that
recovers metric ROV poses and a compact target reconstruction without
external positioning. RGB gating and multi-view orange consensus
stabilise the acoustic measurement; a polar disk residual tolerates
face-versus-centre geometry under an assumed box prior; and a
jump-gated SfM prior absorbs COLMAP teleports without copying them.
On a real freshwater orbit, the estimator reaches centimetre-scale
orange residuals, and cone-arch sonar Gaussian splatting yields a
natively metric cloud near the localised landmark---stronger support
than RGB densification on COLMAP poses under the same views.
Reconstruction remains bounded by 2D FLIS elevation ambiguity and
limited viewpoint diversity; active next-best-view control on the ROV
is the natural next step.

\balance
\bibliographystyle{IEEEtran}

\end{document}